# Clutter Classification Using Deep Learning in Multiple Stages

Ryan Dempsey, Jonathan Ethier
Communications Research Centre Canada (CRC)
Ottawa, Ontario, Canada
ryan.dempsey, jonathan.ethier @ ised-isde.gc.ca

*Abstract*—Path loss prediction for wireless communications is highly dependent on the local environment. Propagation models including clutter information have been shown to significantly increase model accuracy. This paper explores the application of deep learning to satellite imagery to identify environmental clutter types automatically. Recognizing these clutter types has numerous uses, but our main application is to use clutter information to enhance propagation prediction models. Knowing the type of obstruction (tree, building, and further classifications) can improve the prediction accuracy of key propagation metrics such as path loss.

*Keywords—machine learning, satellite imagery, clutter classification, propagation loss*

## I. Introduction

Machine learning-based image classification is a powerful tool with performance that has surpassed heuristic-based classification algorithms [1]. Industries benefit greatly from accurate image classification including applications such as improving inventory management in retail businesses [2], enabling efficient and effective autonomous vehicles in the automotive industry [3], and disease identification in healthcare professions [4]. Wireless communications have similarly been enhanced using machine learning-based classification [5] as well as in regression-style predictions using imagery and convolutional neural network (CNN) technology [6]. This work aims to continue the gainful use of machine learning-based image classification to improve the accuracy of propagation predictions by labeling the type of obstructing clutter in non-line-of-sight (NLOS) communications links.

Typically, propagation models such as the Longley-Rice propagation model [7] account for terrain (ground) information but do not consider the presence of clutter (objects above the ground) in an explicit manner. The consideration of clutter information in propagation models can lead to greater accuracy and more reliable predictions, as is the case in the ITU-R model P.1812 [8]. Modern propagation models like P.1812 can make use of clutter information with a fairly coarse spatial resolution (30 meters) and use representative clutter heights to act as stand-ins for clutter type. This paper works towards a three-fold goal: (1) producing labeled clutter maps with substantially higher spatial resolution, i.e. 1 meter or smaller, (2) improving the labeling of the clutter maps to be more precise representations of various tree and building types, and (3) accomplishing this task in an automated manner, thereby allowing the process to be computationally scalable. Higher-resolution labeled clutter maps are a key requirement for improving machine learning-enhanced propagation models. The fourth longer-term goal of this work will be to generate these more detailed clutter maps for improved training of such models.

The goal of this work is to design a classification system that labels entire satellite images as belonging to one of five classes: coniferous tree, deciduous tree, residential building, non-residential building, and other. These five classes are chosen to strike a balance between high granularity for improving path loss estimation, and computation time. Having a class for each unique species of tree would necessitate an unrealistic number of training examples since we would likely require thousands of samples per unique class when training from scratch [9]. This would require a considerable number of examples and excessive training time for our current model evaluation methodology and our current resources. Similarly, unique classes for individual building materials are also out of the scope of this work.

Having separate classes for deciduous and coniferous trees can significantly improve path loss estimation in two notable ways: (1) the difference in RF attenuation through deciduous and coniferous trees can vary by as much as a factor of 10 and (2) the attenuation through leafy and leafless trees (i.e. deciduous in summer vs. winter) can differ by as much as 20% [10]. Propagation models can exploit this additional information to improve the estimates of RF attenuation through these vegetative structures.

Similarly, the distinction between residential and non-residential buildings can improve path loss estimation over a single building class, simply due to the differences in construction materials [11].

This paper describes a CNN-based method for labeling clutter maps using satellite imagery. The intended use for these labeled clutter maps is to enhance wireless propagation models, but the applications extend beyond the wireless domain. Section II provides an overview of the goals of the work and the classification process used to achieve these goals. Section III describes the datasets and tools used to develop and test our models. Section IV describes the data preparation and the validation process. Section V evaluates the results of the validation, and section VI summarizes our findings.

## II. CLASSIFICATION SYSTEM OVERVIEW

This work involves training convolutional neural networks to classify satellite images as one of the following five classes:

- Deciduous tree
- Coniferous tree
- Residential building
- Non-residential building
- Other

Each image sample is assigned one label to represent the object at the center of the image. We use a two-stage approach, wherein one classifier ("stage 1") is responsible for differentiating higher-level class (tree vs. building vs. other), and another set of classifiers ("stage 2") are used to further classify (i.e. deciduous vs. coniferous and residential vs. non-residential). This method is used rather than a single model because the classifier had lower accuracy when trained to differentiate between all five classes. This was likely because it was trained to learn two different specialized tasks (classifying tree types and classifying building types) as well as the more general task of classifying trees and buildings. The two-stage method makes use of the strength of specialized learners without overwhelming the network. A summary of the classification pipeline can be seen in Fig. 1.

All three models are trained and validated using the same dataset, though, in practice, it should be noted that in cases of mislabeling in stage 1, the sample will be passed to the incorrect model in stage 2. This phenomenon is further explored in section V. The same CNN architecture is used for all three models and is shown in Fig. 2. The original satellite image with a pixel resolution of 640x640 is first center-cropped to 112x112 and then input to the network, which consists of 5 convolutional layers, a max pool layer, and a fully connected layer. Each convolutional layer consists of a 16-channel, 3x3 convolution with same padding, followed by Rectified Linear Unit activation and 2x2 max pooling with a stride size of 2. The final convolutional layer also contains a batch normalization before its pooling. After the convolutional layers, one final max pooling is performed to down-sample the image further, followed by a single dropout at p=0.5. Finally, the fully connected layer classifies the image using a Softmax activation for the single-stage approach and stage 1 of the two-stage approach, or Sigmoid activation for stage 2 of the two-stage approach. Training and validation examples are 8x enhanced with combinations of vertical and horizontal reflections, and 90-degree rotations. Adam optimizer is used to minimize cross-entropy loss. The stage 1 model, each stage 2 model, and the single-stage model have a total of 7875, 7585, and 8165 learnable parameters respectively. This relatively low parameter count emphasizes the modest computational demand imposed by our models.

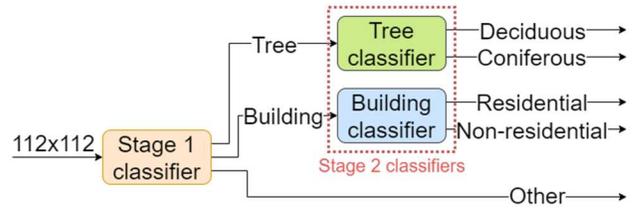

Fig. 1. Proposed two-stage classification pipeline.

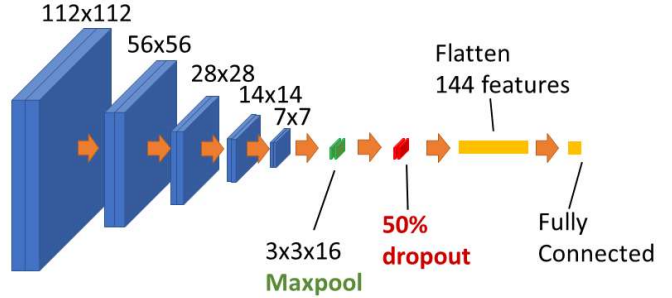

Fig. 2. Convolutional neural network architecture.

## III. RESOURCES

This section lists the datasets and tools used for all training, validation, and testing.

The Ottawa Tree Inventory [12] provides locations and species of trees in the area of Ottawa, Ontario, Canada. The species of trees from the inventory are initially converted into one of two classes: deciduous or coniferous. We draw a random subset of these for cross-validation and testing. The Kingston Tree Inventory [13] is used identically for the independent test case in Kingston, Ontario, Canada.

For building samples, the Open Database of Buildings [14] provides locations of buildings in Canada. This is spatially joined with OpenStreetMap Buildings [15] to determine the building type of each sample. The building types have a variety of classes such as residential, industrial, sports, hospital, etc. We categorize into two building classes: residential and non-residential. Similarly to the tree inventories, we draw a random subset of buildings. Lastly, the "other" class samples are manually generated by sampling open areas without trees nor buildings e.g. fields and parking lots. These three aforementioned datasets are concatenated to create a combined clutter inventory consisting of locations and classes.

To classify the samples, we use satellite images obtained with Google Maps Static API [16]. This provides a 640x640-pixel image at zoom level 20, centered at the location indicated by the entry in the clutter inventory.

The model is implemented in Pytorch and evaluated in sci-kit learn. Additionally, DeepForest [17] [18] is used to ensure the model is trained and evaluated on the correct labels.

## IV. EXPERIMENT METHODOLOGY

To obtain a confidence interval of model performance, k-fold cross-validation is used on the Ottawa data. We use k-means clustering on the sample's coordinates to ensure geographically independent test sets. This gives k clusters of samples that are geographically separated. Each cluster can then be used in one fold as an independent test set where the remaining samples are randomly split between training and validation. In our experiments, this leads to test data in the range of 10-30%, with most test sets being around 20% of the data. Training and validation are always split 80% and 20% respectively.

This method minimizes potential bias from specializing and testing on clutter in the same area, which would likely look the same, or standard randomized cross-validation, which would provide an optimistic estimate of performance [19]. Instead, clustering-based cross-validation evaluates how well a generalized model performs in specific areas without having been influenced by data from those areas [20]. The method also allows us to construct an ensemble model from the k models, which may improve generalization since each fold (area) may have different-looking training data. An ensemble model is created by implementing a majority vote between the trained models from each fold. For a given sample, models "vote" on the class by taking the mode output as the final prediction. This ensemble model is independently tested on data from another city. Kingston is chosen to fairly assess the performance as it is of similar latitude and population density to Ottawa, which would imply similar types of trees and buildings. The same Ottawa data are used for the cross-validation of all models, and the same Kingston data are used for the independent test of all models.

The data cleaning steps we perform before all training, validation, and testing are omitting images with any of the following criteria: (1) discrepancy between an existing tree segmentation model and the listed class, (2) file size below a predetermined threshold, and (3) multiple classes at the same location.

The initial data cleaning involves running an open-source tree segmentation model, DeepForest [17] [18], on all images. This model strictly determines the locations of trees in an image. The following samples are thus discarded from training, validation, and testing:

- Images with a tree in the center 75x75 pixels, with non-tree inventory label
- Images without a tree in the center 75x75 pixels, with tree inventory label

This ensures the models are trained and fairly evaluated on images containing the correct types of clutter (i.e. tree vs. building), though this does not necessarily ensure they are trained on the correct classes (e.g. deciduous vs. coniferous).

Omitting examples below a threshold file size allows only sufficiently high-quality, unambiguous images to be used during training. This size is chosen by investigating at what size the images begin to appear clear to the naked eye. This occurs at 180 KB for our dataset.

Omitting examples with multiple classes at the same location helps ensure that images used for training and testing all have the relevant object unambiguous and unobstructed in the center of the image.

V. RESULTS

A. Model Performance on Test Data

For each model, using the approach outlined in section IV, 5-fold cross-validation is performed. The test set results of all folds are summarized in Table I. The mean accuracy for each class is calculated across all folds, weighted by the number of test samples in the fold.

Given that stage 1 is unlikely to have 100% accuracy, the results in Table I are not a rigorous estimate of the full system's performance on real data. Further analysis can be performed to determine the accuracy of the full classification system for each tree and building class.

TABLE I. INDIVIDUAL STAGE RESULTS OF 5-FOLD CROSS-VALIDATION IN OTTAWA

| Class | Model | Mean accuracy | Standard Deviation |
|---|---|---|---|
| Tree | Stage 1 | **0.992** | 0.00388 |
| Building | Stage 1 | **0.949** | 0.0141 |
| Other | Stage 1 | **0.842** | 0.0999 |
| Deciduous | Stage 1 | **0.992** | 0.00484 |
| Coniferous | Stage 1 | **0.991** | 0.00340 |
| Residential | Stage 1 | **0.970** | 0.0119 |
| Non-residential | Stage 1 | **0.914** | 0.0297 |
| Deciduous | Tree | **0.894** | 0.0380 |
| Coniferous | Tree | **0.803** | 0.0605 |
| Residential | Building | **0.886** | 0.0174 |
| Non-residential | Building | **0.752** | 0.0294 |

Assuming the test performances are an accurate representation of the model's performance across the entire population, let $TPR_i^j$ be the true positive rate of class i, in stage j. For example, $TPR_i^1$ for deciduous trees would be the proportion of truly deciduous trees that were labeled "Tree" in stage 1, and $TPR_i^2$ would be the proportion of truly deciduous trees that were labeled "deciduous" in stage 2. Based on conditional probability paired with our assumption above, it follows that the total TPR of class i is equal to the product of the TPR of each stage. This is shown in (1):

$$TPR_i = \prod_{j=1}^{N} TPR_i^j \qquad (1)$$

In Table II we use these values computed with (1) for the two-stage approach to compare the performance with the single-stage approach, cross-validated on Ottawa data. The standard deviations are calculated using the formula for independent random variables in [21]. The total number of samples used in the cross-validation are as follows: deciduous (5619), coniferous (3390), residential (1559), non-residential (981), and other (2922). Note that these are not the natural class conditionals; additional non-residential buildings were included to balance the cross-validation dataset. The natural class imbalance in both Ottawa and Kingston is such that non-residential samples make up ~10% of buildings.

The two-stage approach is largely an improvement on the single-stage approach though we note that the improvement on the underrepresented non-residential buildings appears to come at the cost of performance on residential buildings.

Following the cross-validations, the two-stage approach is tested as an ensemble model in Kingston to ensure proper generalization on similar cities. Using a majority-voting scheme, the results are summarized in Table III.

TABLE II. COMBINED RESULTS OF 5-FOLD CROSS-VALIDATION IN OTTAWA

| Class | Estimated two-stage accuracy | SD | Single-stage accuracy | SD |
|---|---|---|---|---|
| Deciduous | **0.887** | 0.0379 | **0.858** | 0.0326 |
| Coniferous | **0.796** | 0.0600 | **0.800** | 0.0725 |
| Residential | **0.859** | 0.0199 | **0.865** | 0.0634 |
| Non-res. | **0.687** | 0.0350 | **0.526** | 0.113 |
| Other | **0.842** | 0.0999 | **0.802** | 0.0997 |

TABLE III. INDEPENDENT TEST RESULTS IN KINGSTON

| Class | Model | Accuracy |
|---|---|---|
| Tree | Stage 1 | **0.974** |
| Building | Stage 1 | **0.815** |
| Other | Stage 1 | **0.963** |
| Deciduous | Stage 1 | **0.971** |
| Coniferous | Stage 1 | **0.986** |
| Residential | Stage 1 | **0.828** |
| Non-residential | Stage 1 | **0.694** |
| Deciduous | Tree | **0.819** |
| Coniferous | Tree | **0.754** |
| Residential | Building | **0.815** |
| Non-residential | Building | **0.741** |

The overall performance of the two-stage approach for trees and buildings is calculated in a similar way to the cross-validation. However, since there is no folding, we directly calculate a binary accuracy score (0/1) for each sample. A sample is correct for the two-stage approach if and only if both stages are correct. The precise accuracy for each class can then be calculated directly.

These scores can also be used to perform a one-sided McNemar's test [22], where the alternative hypothesis is that the two-stage method significantly improves accuracy. This is summarized in (2)-(3), where $\mu_2^i$ and $\mu_1^i$ denote the accuracies on class i by the two-stage and single-stage models respectively. For coniferous trees, the alternative hypothesis is reversed, to determine whether the two-stage method is significantly worse. These data are used to make a contingency table for each class, as described in [22]. To determine the p-value for each class, we use a chi-squared test with correction for continuity.

$$H_0: \mu_2^i = \mu_1^i \qquad (2)$$

$$H_A: \mu_2^i > \mu_1^i \qquad (3)$$

The results are summarized in Table IV. The total number of samples used to test on Kingston data are as follows: deciduous (1125), coniferous (280), residential (5852), non-residential (671), and other (898). Contrary to Ottawa, the class conditionals are maintained to fairly test the models.

TABLE IV. COMBINED TEST RESULTS IN KINGSTON

| Class | Two-stage accuracy | Single-stage accuracy | Diff. | p-value |
|---|---|---|---|---|
| Deciduous | **0.790** | **0.748** | 0.0427 | 2.09e-5 |
| Coniferous | **0.746** | **0.782** | -0.0357 | 4.93e-21 |
| Residential | **0.688** | **0.670** | 0.0174 | 4.20e-4 |
| Non-res. | **0.531** | **0.450** | 0.0805 | 8.61e-7 |
| Other | **0.963** | **0.918** | 0.0457 | 2.48e-9 |

Similarly to the single-stage method, the two-stage method also has the lowest classification accuracy on non-residential building samples, primarily mistaking them for residential buildings. This is likely due to the class imbalance on top of the fact that non-residential buildings would likely have a wide variety of appearances. There is still a significant improvement over the single-stage approach. We reject the null hypothesis for all classes except coniferous trees, indicating that performance has increased for these 4 classes with statistical significance. Model classification accuracy on coniferous tree samples decreased with 5% significance (Table IV). Nevertheless, we still suggest the multi-stage approach since it provides >50% sensitivity for all classes as well as improving every non-coniferous class. The f1 score also improved, from 0.692 to 0.719 and 0.585 to 0.605 for micro- and macro-averaged scores, respectively.

B. Imperfect Ground Truth

Qualitatively, our approach has greater success with samples that have very clear boundaries between the dominant object in the image that is to be classified and opposing classes. Successful samples have primarily the same label throughout the center area. This can be seen in Fig. 3, where the residential building in the center (the one being classified) is adjacent to other residential buildings, with only a small tree in the vicinity.

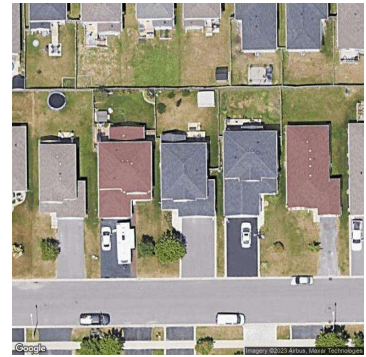

Fig. 3. Successfully classified residential building.

Misclassification also occurs in cases where there are several other objects in the center of the image, causing confusion in the model. An extreme example of this can be seen in Fig. 4, where there are not only several trees in the center of the image, but they are also completely obscuring the actual labeled clutter example: a residential building. Since the image clearly shows abundant foliage, this sample was "mislabeled" as a coniferous tree. This highlights the fact that ground truth training data are not perfect. Despite using tools such as DeepForest to throw away erroneous samples from both testing and training, DeepForest did not successfully detect a tree in the image, allowing it to be tested as a residential building. It is quite possible that DeepForest itself had a similar imperfect ground truth.

This problem affects training as well since the training and validation examples underwent the same filtering process. This makes it more difficult to draw firm conclusions on model performance. If the erroneously labeled samples are incidentally overrepresented in test sets, then the performance is conservatively estimated. However, if the erroneous samples are incidentally overrepresented in training sets, then the model may learn non-ideal associations for classification. This highlights the need for clean, well-labeled data.

This also highlights a related challenge with our approach. By classifying an entire 112x112 image with a single label, errors can creep into the model since the ground truth does not appropriately represent the variety of different types of clutter in the image.

Another issue is the residential and non-residential labels themselves. These labels were chosen as a proxy for building materials to indirectly capture the most significant features relevant to electromagnetic propagation (e.g. wood, drywall, and brick being dominant in residential vs. concrete rebar and metal being dominant in non-residential). At times, however, the label is not sufficiently descriptive of the building materials.

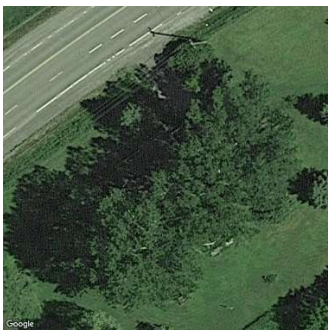

Fig. 4. Cluster of trees hiding a residential building.

Fig. 5 shows a non-residential building that the model labeled as residential. It has roof material and appearance similar to typical residential buildings, which is likely why the model mislabeled it – a human would potentially make the same mistake. Given that examples of this kind may be present in the training set as well, this highlights an issue with our labeling system – the residential and non-residential classes may not always consistently capture the material properties, which would be ideal for wireless propagation modeling; and there are no guarantees of what visual relationships the model may learn to distinguish between the two building classes. In this case, the model successfully captured the relevant information, but this may not be consistent with all samples.

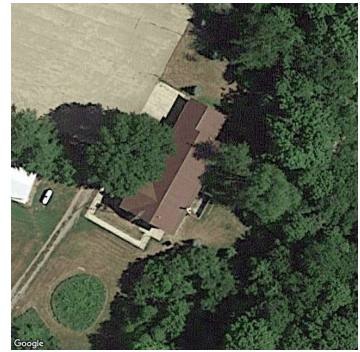

Fig. 5. Non-residential building classified by model as residential.

### C. Future Recommendations

The main issues from section V.B can be summarized as follows: difficulty predicting with more complicated images with varied clutter, and building labels not always being indicative of building material (and therefore providing unhelpful information to a future propagation model).

To solve the first issue, we suggest using a segmentation model for stage 1. This could leverage existing semantic segmentation models such as [23] for transfer learning. This would allow pixel-wise classification into trees and buildings, with further classification of the individual objects by a classifier, taking advantage of both the precision classification of segmentation and the specialized classification of CNNs.

To solve the building problem, future work can include more precise building classes that are more directly related to materials (e.g. glass, brick, wood, other). This would (1) provide finer-grain resolution and (2) ensure the classes are correlated with materials and therefore propagation. As with most machine learning problems, a significant portion of the difficulty would be in finding accurate and precise labels. OpenStreetMap Buildings does contain additional info such as height and the number of floors, however, the data is scarce. Further work could also explore the use of street-level views of buildings to better classify the type of building. A classifier that has access to both satellite imagery and street-level imagery could lead to improved building (and tree) classification. The work in [24] classifies residential and non-residential buildings using Google Streetview and a similar neural network architecture with strong results. However, no mention is made of train-test split nor per-class sensitivities. Duplicating our experimental methods while combining multiple image angles would likely improve classification by reducing the ambiguity caused by rooftop structures.

One benefit to the multi-stage classification system is that the "other" class could also be further classified by more stage 2 models for various applications, e.g. field vs. parking lot vs. other. Similarly, a third stage could be added to further classify deciduous and coniferous trees e.g. maple vs. non-maple and pine vs. non-pine. This flexibility allows further refinement if propagation models benefit from it, without needing to adjust the original models.

An additional source of improvement may come in the training process. This work involved an ensemble model trained in Ottawa and tested in Kingston. Within the cross-validation, naturally, the test results outperformed the independent Kingston test, since it was trained in that city. An alternative approach could be to train a model for different regions in Canada, allowing the model to learn more specific details about the look of the region's clutter. If this approach is used, geographic cross-validation should be performed to prevent an exaggerated estimate of performance.

## VI. Conclusions

This paper discussed the use of convolutional neural networks for classifying buildings and foliage. Both the two-stage and single-stage classifications provide strong performance when classifying tree types, but the single-stage classifier frequently mislabels non-residential buildings as residential buildings. Separating the classification into two stages can improve performance only if the first stage has nearly perfect performance. Feeding an imperfect system to another imperfect system may not always work effectively due to the stacking of conditional probabilities. We have shown that our methodology does train a near-perfect stage 1 classifier (building vs. tree being a fairly "easy" classification), allowing the second stage models to specialize entirely on either trees or buildings and resulting in a significant improvement over the single-stage approach on all but one class. This work demonstrates a reliable method to automate clutter labeling in Canada, which can be used to enhance propagation models.

Future work to improve this process includes replacing the high-level first stage with a semantic segmentation model and classifying each object, as well as developing more specific classes in additional stages. Having dedicated models for unique regions in Canada may improve performance by learning more specific features of a region's clutter.